# Artificial Image Tampering Distorts Spatial Distribution of Texture Landmarks and Quality Characteristics

Tahir Hassan
*School of Computing*
*The University of Buckingham*
Buckingham, UK
1405862@buckingham.ac.uk

Aras Asaad
*School of Computing*
*The University of Buckingham*
Buckingham, UK
aras.asaad @buckingham.ac.uk

Dashti Ali
*Independent Researcher*
Ontario, Canada
dashti.a.ali@gmail.com

Sabah Jassim
*School of Computing*
*The University of Buckingham*
Buckingham, UK
sabah.jassim @buckingham.ac.uk

*Abstract*—Advances in AI-based computer vision has led to a significant growth in synthetic image generation and artificial image tampering with serious implications for unethical exploitations that undermine person's identification and could make render AI predictions less explainable. Morphing, Deepfake and other artificial generation of face photographs undermine the reliability of face biometrics authentication using different electronic ID documents. Morphed face photographs on e-passports can fool automated border control systems and human guards. This paper extends our previous work on using the persistent homology (PH) of texture landmarks to detect morphing attacks. We demonstrate that artificial image tampering distorts the spatial distribution of texture landmarks (i.e., their PH) as well as that of a set of image quality characteristics. We shall demonstrate that the tamper caused distortion of these 2 slim feature vectors provide significant potentials for building explainable (Handcrafted) tamper detectors with low error rates and suitable for implementation on constrained devices.

## I. INTRODUCTION

Morphing attacks on face biometrics were reported first in 2014 by Ferrara et al. [1] and have been perceived to be difficult to detect. It is based on image fusion techniques that on the input of two (or more) genuine aligned face images produce morphed face images similar to the input face images [1], see Fig. 1. Electronic Machine-Readable Travel Documents (eMRTD) that contain genuine face biometric reference images of the e-passports holder are increasingly deployed worldwide. To obtain a genuine eMRTD, one needs to provide a standard digital or printed passport image in person (or online) to the issuance authority. Printed passport images will be scanned to obtain a digitized eMRTD. Finally, the digital or print-scanned photograph will be stored in eMRTD during the production process. Morphing attacks can happen at different stages during the process of applying and issuing eMRTD. Submitting a morphed photograph maliciously by an applicant can be used to utilize intrinsic intra-class variation tolerance by face recognition systems (FRS), resulting in serious vulnerability [2]. Morphed faces can bypass human experts as well as commercial FRS schemes deployed in border controls [3]. Criminals aim to morph their face photo with someone who is not blacklisted by police to gain access to protected countries or territories.

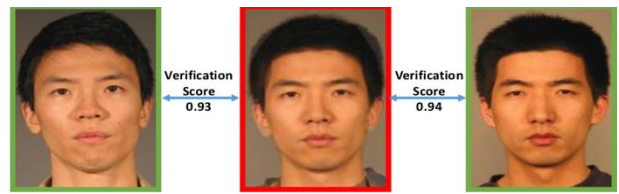

Figure 1. An example of Commercial off-the-shelf FRS verification score operating @ 0.1% false acceptance rate obtained from comparing a morphed face with corresponding genuine faces [2].

Hence, preventing/detecting such attacks is crucial to maintain the integrity of FRS in airports and similar applications where FRS is deployed. In Fig. 1, we show an example of a verification score obtained by a commercial FRS when comparing two genuine faces with their morphed version.

To prevent morphed faces from passing through FRS, several morph detection algorithms have been proposed in the last 5 years by the biometric and forensics community. Raghavendra et al [4] proposed the use of local micro-textures based on Binary Statistical Image Features (BSIF) and local binary patterns (LBP) for the purpose of morph detection. Makrushin et al [5] used Benford's law based on quantized Discrete Cosine Transform coefficients to differentiate morphs from genuine faces. Mathematical field of Topological Data Analysis (TDA) and its main tool of Persistent Homology (PH) explored by a series of papers for the purpose of fake face detection [6]–[8]. Many Deep Learning (DL) approaches used to detect morph in both transfer learning setting and from scratch [9]–[11]. Hildebrandt et al [12] used the stir-trace approach to detect morphed images, and image degradation approach explored by Neubert [13] where the focus is on corner feature detectors and its effect for morph detection. We refer interested readers to [14] for a comprehensive survey of morph detection and generation algorithms.

Another approach for morph detection is based on image quality measurements. The high perceptual quality of morphed face images by any morph technique is vital to guarantee the success of morphing attacks; otherwise, any visible artifact can be detected by human operators. Reference-based image quality measurements like peak signal-to-noise ratio (PSNR) and structural similarity index measure (SSIM) cannot efficiently classify bonafide images

The first author would like to thank Ten-D Innovations for sponsoring his research.

from morphed ones being applied on the same morphing pair [15]. Recently, Fu et al [16] conducted an interesting extensive study on the effect of face image morphing on face image quality, they investigated twelve different image quality metrics divided into two groups G1: FaceQnet [17], rankIQ [18], Magface [19], SER-FIQ [20]; and G2: BRISQUE [21], NIQE [22], PIQUE [23], CNNIQA [24], DeepIQA [25], MEON [26], dipIQ [27], RankIQA [28]. They conducted experiments on digital and re-digitized images and experimentally demonstrated that most of these quality metrics fail to detect the image artifacts produced by morphing techniques and, therefore, are unsuitable for morph detection [16]. MagFace was the only technique that efficiently separated bonafide images from morph images. Moreover, they found that MEON achieves high performance in separating genuine images from morph ones when applied on certain areas such as tightly cropped face, nose, and eyes.

Our previously proposed PH landmark-based morph detector, see [7], [8], exploit the fact that the seemingly invisible distortion caused by morphing result in spatial displacement of texture landmarks in the genuine participant images. The work in [16] provides further evidence to the perception that image quality, as well as the effect of morphing are not uniformly distributed through different image regions. It also, shows that single quality metrics have their limitations in dealing with morphing attacks.

The current work is based on exploiting, the above mentioned, dual impact of morphing attacks on spatial distribution of texture landmarks and the scatter of multiple image quality characteristics. We shall design an innovative no-reference image quality feature vector, the coordinates of which encapsulate the scatter of a variety of image quality metrics/characteristics mimicking Human Vision System (HVS) quality characteristics. We shall also complement our previously proposed PH morph detector, by introducing a new simple approach for classification of landmark-based PH barcodes. We then conduct experimental work on two face-image databases to show how even with such small-size feature vectors one can build efficient high performing automatic morph detectors.

## II. CONTRIBUTION OF THE PAPER

Naturally, human experts examining face photos for detection of foul play, anticipates that morphed images are of slightly degraded quality and thus scan the submitted image for the presence of artefacts, blurring and/or minutely distorted facial regions. The main challenge in automating this process, is relating to quantifying the sought after distortion/ degradation measures in a very efficient manner. The main contributions of this manuscript that attempt to address this challenge include:

### A. Multi-Characteristics Image Quality feature (MCIQ)

There are many ways of characterizing image quality including level of blurriness, distortion, illumination, color distortion, etc. Most existing image quality measures require the use of reference images. In section III, below, we introduce an innovative reference-less spatio-statistical image quality feature vector representing the scatter of various quality characteristics in the image. This feature vector was recently designed, by the first author, to compare tumor ultrasound scan images generated by different clinical centers in relation to overfitting of DL models of tumor classification. Her we use it as a morph discriminating tool.

### B. Statistical parameters of PH Barcodes

We propose a novel PH-based algorithm to detect morphed faces which rely on extracting statistics and persistence Betti curves from the space of persistence barcodes. For efficiency purposes, topology construction is restricted to the $4^{th}$ rotation of the uniform local binary geometry patterns (ULBP) $G_4$. In other words, we use one rotation and one landmark geometry instead of our previous more computationally demanding topology constructions that were based on 8 rotations.

### C. Generalization Tests (Cross-databases validation)

Generalize the strength of the two proposed methods by conducting validation across different databases (DB), whereby we train SVM classifier on one image DB and test it on a different DB and achieve less than 4% in terms of false rejection rate (FRR) and false acceptance rate (FAR).

## III. THE MCIQ IMAGE FEATURE VECTOR

In this section, we describe the structure of our novel MCIQ spatio-statistical feature vector that combine several image quality metrics to detect face image morphing attacks. Unlike reference-based image quality metrics (e.g., PSNR and SSIM), MCIQ is self-reference and is independent of image size. It is inspired by the Universal Image Quality Index (UIQI) which combines three quality characterizing factors: loss of correlation, luminance distortion, and contrast distortion in order to measure image distortion with respect to a given reference [29]. Then, UIQI was modified by adding one more factor called modified skewness to the other three components, which is shown to perform better [30]. Here, instead of finding an image quality index (one value index), we construct a multiple histogram-based feature vector that measures distortion between image blocks of an input image in terms of various metrics including Correlation, Luminance, Contrast, Kurtosis, and Skewness. These five metrics together form a powerful feature vector suitable for morph detection and classification tasks that are image degradation related.

To define the five components of MCIQ, first, we state the five related basic statistical parameters. Let $x = \{x_1, x_2, \cdots, x_n\}$ and $y = \{y_1, y_2, \cdots, y_n\}$ be two given real-value gray-pixel sequences representing two equal size image blocks, then:

$$\bar{x} = \frac{1}{n}\sum_{i=1}^{n} x_i \qquad (1)$$

$$\sigma_x^2 = \frac{1}{n-1}\sum_{i=1}^{n}(x_i - \bar{x})^2 \qquad (2)$$

$$\sigma_{xy} = \frac{1}{n-1}\sum_{i=1}^{n}(x_i - \bar{x})(y_i - \bar{y}) \qquad (3)$$

$$s_x = \frac{\sum_{i=1}^{n}(x_i - \bar{x})^3}{(n-1)(\sigma_x)^3} \qquad (4)$$

$$k_x = \frac{\sum_{i=1}^{n}(x_i - \bar{x})^4}{(n-1)(\sigma_x)^4} \qquad (5)$$

Where $\bar{x}$ represents the mean of $x$, $\sigma_x^2$ represent its variance, $\sigma_x = \sqrt{\sigma_x^2}$ is its standard deviation, $\sigma_{xy}$ stands for the covariance of $x$ and $y$, $s_x$ and $k_x$ represent the skewness and kurtosis of $x$ respectively.

The MCIQ is constructed in steps. Starts by partitioning the input image (a gray-scale image) into 36 same-size rectangular blocks, see Fig. 2. Next compute the quality indices of each image block with respect to the other 35 blocks in terms of the 5 image quality indices (Correlation, Luminance, Contrast, Kurtosis, and Skewness) using the following formulae:

$$Correlation_{xy} = \frac{\sigma_{xy}}{\sigma_x \sigma_y} \qquad (6)$$

$$Luminance_{xy} = \frac{2\bar{x}\bar{y}}{(\bar{x}^2 + \bar{y}^2)} \qquad (7)$$

$$Contrast_{xy} = \frac{2\sigma_x \sigma_y}{(\sigma_x^2 + \sigma_y^2)} \qquad (8)$$

$$Kurtosis_{xy} = \frac{2k_x k_y}{(k_x^2 + k_y^2)} \qquad (9)$$

$$Skewness_{xy} = \frac{2s_x s_y}{(s_x^2 + s_y^2)} \qquad (10)$$

The computed indices for each of the 5 quality indices can be organized in a $36 \times 36$ matrix which is symmetric with unit diagonal. The 630 values, which is $(36 \times 35)/2$, above the diagonal represent the distribution of the related quality between block pairs of the partitioned input image. We then quantize these 630 indices into 10 equal bins. In other words, the relation between the image-blocks for each quality measure is represented by a histogram vector of length 10. Finally, the 50-dimensional MCIQ feature vector is constructed with equal contributions from the selected image quality characteristics in the order: [Correlation, Luminance, Contrast, Kurtosis, Skewness]. Fig. 3, below, displays MCIQ features for a genuine image (A) and a morphed one (B), where one can notice differences between the distortion in the histograms of each quality characteristic.

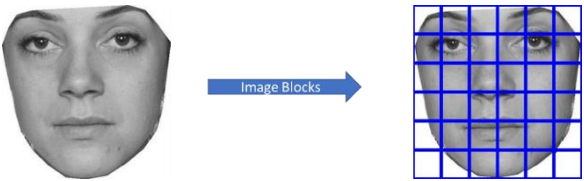

Figure 2. Partition of a face image into 36 equal rectangular blocks.

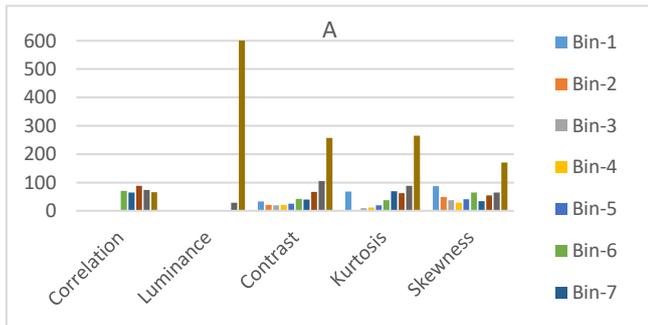

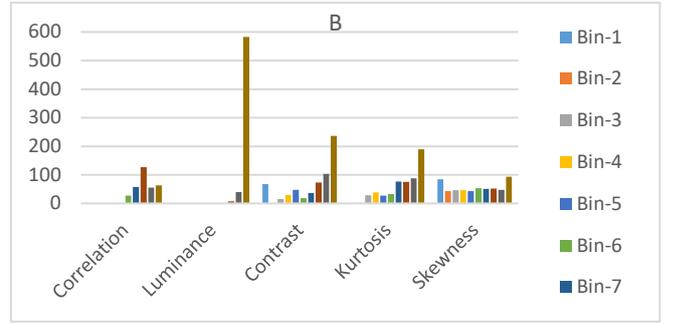

Figure 3. MCIQ feature vectors of a genuine image (A) and a morphed image (B) from AMSL DB.

IV. PERSISTENT HOMOLOGY

Persistent Homology (PH) is one of the main tools in the field of TDA which relies on extracting birth and death of homological invariants during a filtration of simplicial complexes built from point clouds. In this section, we describe the mathematical background necessary to define PH and filtration process from image pixel landmarks that can be computed automatically.

*A. Vietoris-Rips and Filtration of Simplicial Complexes*

Mathematical elements analyzed by PH are known as simplices, being the building blocks of higher-dimensional counterpart graphs called simplicial complexes (SC). An $n$-simplex is defined as a convex hull of an $n + 1$ affinely independent points. A 0-simplex is a vertex, 1-simplex is an edge, 2-simplex is a filled triangle, and a 3-simplex is a solid tetrahedron, see Fig. 4. A finite collection of such simplices forms a finite simplicial complex which satisfies some conditions [31]. In this work, we use the Vietoris-Rips (VR) algorithm to construct SCs associated with image pixel landmarks selected for their texture geometric semantics.

Let $P$ be a point cloud in $\mathbb{R}^2$, $\epsilon$ be any non-negative real number, and $VR(P, \epsilon)$ be the Vietoris-Rips SC formed by connecting all pairs of points in $P$ for which the Euclidean distance between them is $\leq \epsilon$. An ordered set $\{p_1, p_2, ..., p_n\}$ spans an $n$-simplex if and only if $(p_i, p_j) \leq \epsilon$, $for\ 0 \leq i, j \leq n$. Increasing $\epsilon$ results in growing the simplicial complex of the given point cloud. This process defines a filtration of simplicial complex, which is a process of building a nested sequence of $VR(P, \epsilon_1) \subseteq VR(P, \epsilon_2) \subseteq \cdots \subseteq VR(P, \epsilon_t)$, where $\epsilon_1 \leq \epsilon_2 \leq \cdots \leq \epsilon_t$. Here, topological features appear and disappear during the filtration that are stored as bars in what is known as persistence barcode (PB). We direct the interested reader to [31], [32] for more details about the mathematics of PH.

*B. Uniform Local Binary Patterns (ULBP)*

Our TDA approach for image analysis is based on selecting point clouds that represent meaningful geometric texture and are automatically determined from image data. There are many such types of landmarks, but we use the local binary patterns (LBP) transform that provide an easy way to automatically determine texture descriptor. It was proposed more than two decades ago and successfully deployed in many computer vision applications [8]. It works by scanning 3x3 patches of an image and encodes each pixel with an 8-bit binary code by comparing central pixel with its 8-neighbours in a clockwise orientation starting from top left corner. By ULBP, we refer to LBP codes that have either 0 or 2 circular-transitions between 0's and 1's. There are 58 ULBP codes which can be grouped into 7 classes according to the number of 1's in their binary codes excluding 00000000 and 11111111. For example, a ULBP code with 4 ones together with its corresponding rotations is denoted as $G_4$-ULBP, see Fig. 5.

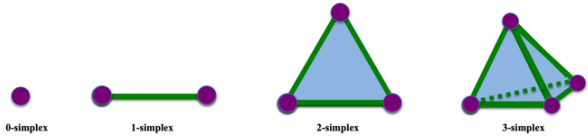

Figure 4. Simplices of dimension 0, 1, 2 and 3.

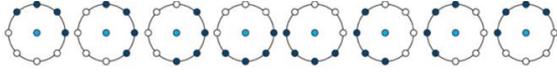

Figure 5. Geometric representation of G4-ULBP codes and its 8 corresponding rotations.

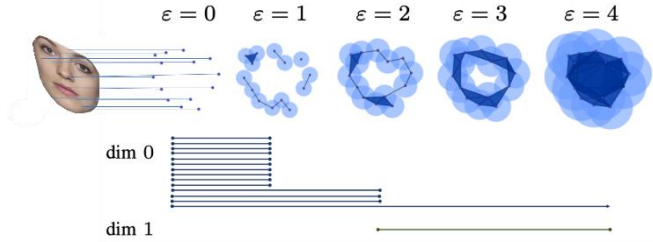

Figure 6. PH construction pipeline. (a) 0-simple selection using ULBP and filtration process, (b) its persistence barcode in dimension 0 (connected components) and 1 (loops/holes).

Point cloud set $P$, which is the building block of PH pipeline, is the input image central pixel locations of the 4$^{th}$ rotation of $G_4$-ULBP, i.e., 01111000. The reason behind choosing $G_4$-ULBP is backed by the results reported in [8] where they concluded that $G_4$-ULBP is one of the best performing geometries after concatenating topological features of all 8 rotations. Here, we opt to select a single rotation from $G_4$-ULBP, 4$^{th}$ rotation and report its performance for morphed face detection. There is no particular reason behind selecting the 4$^{th}$ rotation, and one can select other rotations of $G_4$-ULBP. The construction of the Vietoris-Rips SC and its filtration for our set of landmarks in a face image is illustrated in Fig. 6.

### C. Persistence Barcode Featurisation

Persistent topological features stored in barcodes are amenable to many machine-learning and statistical tasks. Hence, we transform (i.e., featurize) the space of PB using two approaches: barcode-binning (i.e., Betti curves) and barcode statistics. Below briefly describe them.

*1) Barcode-Binning (BB):* This approach is one of the simple vectorization methods that relies on counting the number of bars in persistence barcodes that intersect with each vertical line $V = 0, 1, 2, ..., \omega$. In this paper, we set $\omega = 24$ equidistance vertical lines. Thus, a topological feature vector of size 25 was obtained for different dimensions of PBs. This feature vector is similar to persistence Betti curves, except that we terminate counting Betti numbers at a specific threshold $\omega$.

We denote the count of bars in dimensions zero and one by BB_D0 and BB_D1, respectively.

*2) Barcode Statistics(BS):* The simplest approach to vectorize the space of PB is to extract statistics directly from PB. In this work, we collect only 10 statistical measurements: average and standard deviation of birth, death and lifespan of bars, median of birth, death, and lifespan of bars and finally, the number of bars. Statistics of birth of topological features in dimension zero of PBs are zero by default and can be ignored. Statistics of PB in dimension zero and one is denoted by BS_D0 and BS_D1, respectively.

### V. DATABASES AND EVALUATION METRICS

To evaluate the performance of the proposed 3 morph detection schemes (MCIQ, BB, and BS), two face image databases (DB) are used for the training and testing experiments. These are the AMSL DB which is available online freely upon request [13], and the Utrecht DB [3]. Morphs are generated using an in-house method by AMSL group in Magdeburg University. For both databases, morphed faces are generated using Combined approach developed in [13] which is a landmark-based approach and produces visually faultless morphed faces. There are 102 genuine faces in AMSL DB and 67 genuine faces in the Utrecht DB. For each of the two DBs, we generated 1000 morphed images from randomly selected genuine face image pairs in the corresponding DBs. Fig. 7, below, displays samples of cropped genuine and morphed face images from the 2 DBs. Cubic SVM is utilized for training and testing under the 5-fold-cross-validation (5-FCV) setting to differentiate genuine faces from their fake counterparts.

Due to the imbalance nature of both databases, we randomly select a sample of 102 and 67 morph images from AMSL and Utrecht, respectively, 10 times, and report average (Avg) and standard deviation (Stdev) of False Rejection Rate (FRR) and False Acceptance Rate (FAR) in the 5-FCV. FRR is the rate of falsely rejected genuine faces, while FAR is the likelihood of falsely accepting a fake image by the classifier. In practice, we aim to minimize both FRR and FAR. The lower the values of FRR and FAR, the more robust the face biometric authentication system is against morphing attacks.

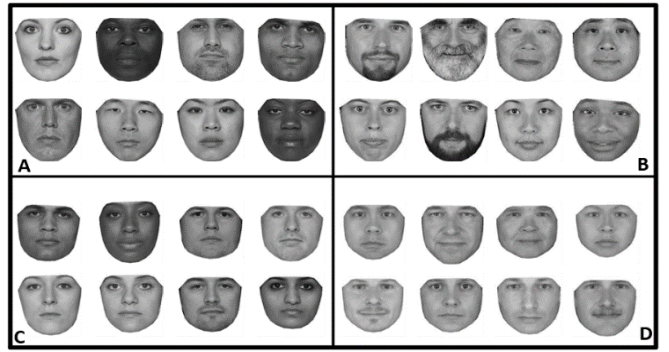

Figure 7. Cropped Face image samples: (A) Genuine images from AMSL (B) Genuine images from Utrecht DB (C) morphed images from AMSL (D) Morphed images from Utrecht DB

### VI. EXPERIMENTAL RESULTS AND DISCUSSION

Experimental results on both AMSL and Utrecht DBs using Cubic SVM are illustrated in Table 1 and 2, respectively. For each feature type, we report 5FCV - (Avg and Stdev) of FRR and FAR in percentages. In these tables best performing results obtained from each DB are highlighted in bold. These results confirm beyond any doubts about the viability of using any of these different features for detecting morphing attacks with high accuracy. However, dim 1 PH-based feature vectors are less successful than the other features. Comparing the performance of the MCIQ feature vector over the two databases shows that it achieves lower FRR when tested with the AMSL images compared with its FRR for the Utrecht images. The Stdev of the FRR over the Utrecht DB exceeds that over the AMSL by ~4%. The FAR rates achieved by the MCIQ method are almost similar in both DBs.

Comparing performance of the PH-based features over the two databases yields a reverse pattern, i.e., their performances over the Utrecht DB is better than that over the AMSL DB. For both DBs, dim 0 PH features outperform the dim 1 PH-features, but with barcode statistics (BS) for AMSL DB and barcode binning (BB) for Utrecht DB. For the AMSL DB, we observe that the BS statistics of the persistence barcodes perform better in detecting morphs while persistent Betti curve (i.e. binning) works better in Utrecht DB especially in terms of FAR. These results demonstrate the success of using PH features constructed from ULBP landmarks in detecting the effect of morphing on the spatial distribution of these texture landmarks. In non-TDA image analysis schemes in the LBP domain rely on histograms of ULBP landmarks. In [8], the difference between the percentage of the ULBP groups in genuine and morphed images from the two DBs were so marginal, and it would not be possible to use these statistics to detect morphing attacks.

It is not easy to explain the differences in the pattern of performances of the two different types of features over the two DBs. However, the two DBs differ in terms of gender, ethnic and skin-color diversity of participants whereby the AMSL is more diverse. These differences may impact the discriminability of the different features. Since MCIQ represents the scatter of various quality characteristics may explain why it is performing better on AMSL.

In our previous work [6]–[8], we tested the performance of the BB PH features, with large size vectors, on these two DBs. The results of the current experiments with PH outperform all previous results of using BB PH features. This can be attributed to two factors: (1) in the current experiments we segmented and cropped the frontal faces and resized to a fixed size whereby the issue of distance from the camera resolved which exists in unprocessed AMSL images.

Table 1. SVM classification errors (%) for AMSL DB.

| Error | Stats. | MCIQ | BB_D0 | BB_D1 | BS_D0 | BS_D1 |
|---|---|---|---|---|---|---|
| FRR | Avg | **0.77** | 3.23 | 9.75 | 2.23 | 5.88 |
| | Stdev | **1.56** | 3.57 | 6.20 | 3.16 | 4.70 |
| FAR | Avg | 2.07 | 3.04 | 10.79 | **1.79** | 7.25 |
| | Stdev | **3.38** | 3.41 | 6.70 | 3.36 | 4.61 |

Table 2. SVM classification errors (%) for Utrecht DB.

| Error | Stats. | MCIQ | BB_D0 | BB_D1 | BS_D0 | BS_D1 |
|---|---|---|---|---|---|---|
| FRR | Avg | 4.18 | **1.67** | 8.36 | 1.87 | 4.08 |
| | Stdev | 5.60 | **3.22** | 5.49 | 2.85 | 6.23 |
| FAR | Avg | 2.21 | **0.30** | 4.16 | 1.65 | 5.82 |
| | Stdev | 3.57 | **0.66** | 5.44 | 3.08 | 6.14 |

Table 3. Generalization tests of both MCIQ and PH features by training on AMSL and testing on Utrecht DB and vice versa.

| Features | Training | Testing | Stats | FRR | FAR |
|---|---|---|---|---|---|
| MCIQ | Utrecht | AMSL | Avg | 1.96 | 24.51 |
| | | | Stdev | --- | 3.52 |
| | AMSL | Utrecht | Avg | 2.99 | 2.99 |
| | | | Stdev | --- | 2.11 |
| BB-D0 | Utrecht | AMSL | Avg | **3.92** | **2.16** |
| | | | Stdev | --- | **1.59** |
| | AMSL | Utrecht | Avg | **2.99** | **1.49** |
| | | | Stdev | --- | **1.41** |
| BS-D0 | Utrecht | AMSL | Avg | 10.78 | 4.51 |
| | | | Stdev | --- | 2.13 |
| | AMSL | Utrecht | Avg | 1.49 | 1.49 |
| | | | Stdev | --- | 1.41 |

(2) Beyond Betti curve, extracting statistics from persistence barcodes helped to boost the performance in AMSL DB while maintaining the same performance for Utrecht DB.

To test the generalizability power of the features proposed in this paper, we opt to choose the best performing features across both DBs which are MCIQ, BB_D0 and BS_D0. In Table 3, we show the results obtained by training SVM using MCIQ and PH features on AMSL DB and test the trained SVM model on Utrecht DB and vice versa. Specifically, we select the best performing SVM model in the 5FCV while trained on AMSL (102 genuine faces with 102 morphed faces subsampled from 1000 morphed faces and repeated 10 times) then used it in testing phase on Utrecht. In the testing phase, we repeated the random subsampling of morphed images due to the highly imbalance nature of both DBs, hence we report the (Avg and Stdev) - FAR and (Avg) – FRR of the 10 subsamples in testing phase. There is no Stdev – FRR as the genuine images are the same in the 10 repeated times. The results are shown in Table 3.

VII. REPRODUCIBILITY AND IMPLEMENTATION DETAILS

Topological features were built and extracted using our DAAR software [33] which is based on: ULBP and Ripser to build Vietoris-Rips filtration. SVM classification is performed in MATLAB (V. 2021b) using cubic kernel and all other parameters left as default. Frontal face images were segmented using dlib-ml [34] first, then cropped out and resized to 280×270 (Height × Width) before feeding to DAAR. The MCIQ feature vector components are all computed using either built-in functions in MATLAB (V. 2021b) or following the settings described in section III above.

VIII. CONCLUSION

We presented two types of novel slim efficiently extracted image feature vectors and incorporated those features into morph detection algorithms that performed well when trained and tested on two relevant Face image DBs and maintained reasonably high performance under generalized cross DB-testing. Both feature vectors reflect spatial distribution of texture landmarks and image quality characteristics. The first uses the TDA tool of PH on texture landmark and the second is the spatio-statistical feature MCIQ, whose components are closely associated with HVS known quality measures. All these efficiently extracted feature vectors are of relatively low dimensionality that are not only suitable for implementation on constrained devices but unlike many machine learning algorithms, its decisions are explainable by adding visualization facilities. Analysis of the MCIQ features and their associated matrices can be used to detect specific regions that their various quality characteristics is suspected of being degraded, relative to the remaining regions, as a result of possible morphing attacks. Furthermore, the spatial distribution of the ULBP can be visualized to detect possible distortion.

One limitation of this work is related to the limited amount of testing/training experiments DBs due to the lack of publicly available large scale DBs of face photos standardized for inclusion in ID-documents (e.g., e-passports).

The performance testing of the PH-based features in this paper was confined to a single ULBP landmark set, and it is essential to extend these experiments in the future for other ULBP landmark sets. Furthermore, future research directions need to focus on other TDA considerations including different filtration descriptors as well as barcode featurization beyond Betti curves and their statistics. The performance of MCIQ feature vector can be improved by investigating similarly structured quality feature vectors but computed in the gradient image or other transformed versions.